%% file: main.tex
\documentclass[letterpaper, 10 pt, conference]{ieeeconf}  

\IEEEoverridecommandlockouts                              

\usepackage{graphics} 
\usepackage{graphicx}
\usepackage{times} 
\usepackage{amsmath} 
\usepackage{amssymb}  
\usepackage{mathrsfs}
\usepackage{amsfonts}
\usepackage{kantlipsum}

\makeatletter
\let\NAT@parse\undefined
\makeatother
\usepackage[numbers,sort&compress]{natbib}

\usepackage{hyperref}
\usepackage{graphicx}
\usepackage{float}
\usepackage{ctable}
\usepackage{cuted}
\usepackage{colortbl}
\usepackage{multirow}
\let\labelindent\relax
\usepackage{enumitem}

\usepackage{amsmath}
\usepackage{algpseudocode}
\usepackage{algorithm}

\usepackage{graphicx}
\usepackage{float}
\usepackage{ctable}
\usepackage{cuted}
\usepackage{colortbl}
\usepackage{multirow}
\usepackage{threeparttable}
\usepackage{makecell}
\usepackage{wrapfig}

\title{\LARGE \bf
GAMMA: Generalizable Articulation Modeling and \\ Manipulation  for Articulated Objects
}

\author{Qiaojun Yu$^{1}$, Junbo Wang$^{1}$, Wenhai Liu$^{1}$, Ce Hao$^{2}$, Liu Liu$^{3}$, Lin Shao$^{2}$, Weiming Wang$^{4}$ and Cewu Lu$^{1*}$
\thanks{$^{1}$Qiaojun Yu, Junbo Wang, Wenhai Liu, Cewu Lu are with Department of Computer Science, Shanghai Jiao Tong University, China. *Cewu Lu is the corresponding author. \texttt{\{yqjllxs, sjtuwjb3589635689, sjtu-wenhai, lucewu\}@sjtu.edu.cn}}
\thanks{$^2$Ce Hao, Lin Shao are with Department of Computer Science, National University of Singapore, Singapore, \texttt{cehao@u.nus.edu, linshao@nus.edu.sg}}
\thanks{$^3$Liu Liu is with Department of Computer Science and Information Engineering, Hefei University of Technology, China, \texttt{liuliu@hfut.edu.cn}}
\thanks{$^4$Weiming Wang is with Department of Mechanical Engineering, Shanghai Jiao Tong University, China, \texttt{ wangweiming@sjtu.edu.cn}}
}

\begin{document}

\maketitle
\thispagestyle{empty}
\pagestyle{empty}

\input{text/abs}

\input{text/introduction}
\input{text/related_work}

\input{text/problem}

\input{text/method}

\input{text/experiment}
\input{text/conclusion}

\input{text/acknowledgments}

{\small
\bibliographystyle{IEEEtranN}
\bibliography{ref}
}

\end{document}

%% file: text/abs.tex
\begin{abstract}
Articulated objects like cabinets and doors are widespread in daily life. However, directly manipulating 3D articulated objects is challenging because they have diverse geometrical shapes, semantic categories, and kinetic constraints. 
Prior works mostly focused on recognizing and manipulating articulated objects with specific joint types. They can either estimate the joint parameters or distinguish suitable grasp poses to facilitate trajectory planning. Although these approaches have succeeded in certain types of articulated objects, they lack generalizability to unseen objects, which significantly impedes their application in broader scenarios. 
In this paper, we propose a novel framework of Generalizable Articulation Modeling and Manipulating for Articulated Objects (GAMMA), which learns both articulation modeling and grasp pose affordance from diverse articulated objects with different categories. In addition, GAMMA adopts adaptive manipulation to iteratively reduce the modeling errors and enhance manipulation performance. We train GAMMA with the PartNet-Mobility dataset and evaluate with comprehensive experiments in SAPIEN simulation and real-world Franka robot. Results show that GAMMA significantly outperforms SOTA articulation modeling and manipulation algorithms in unseen and cross-category articulated objects. Images, videos and codes are published on the project website at: ~\href{http://sites.google.com/view/gamma-articulation}{\texttt{sites.google.com/view/gamma-articulation}}.
\end{abstract}


%% file: text/introduction.tex
\section{Introduction}

Articulated structures like doors and drawers are commonly used in daily environments to store objects and divide spaces. They comprise interconnected parts governed by specialized joints, such as revolute and prismatic joints, which constrain the degrees of freedom according to their kinematic structure\cite{liu2022toward, li2020category}. 
When humans manipulate articulated objects, we can recognize the physical structure like a pivot with our prior knowledge and easily apply such skills to unseen circumstances.
Likewise, artificial intelligence (AI)-empowered robots also show the potential to mimic humans' intuition and actively interact with articulated objects in awareness of kinetic constraints\cite{gadre2021act, jiang2022ditto}.

\begin{figure}[th]
    \centering
    \includegraphics[width=0.8\columnwidth]{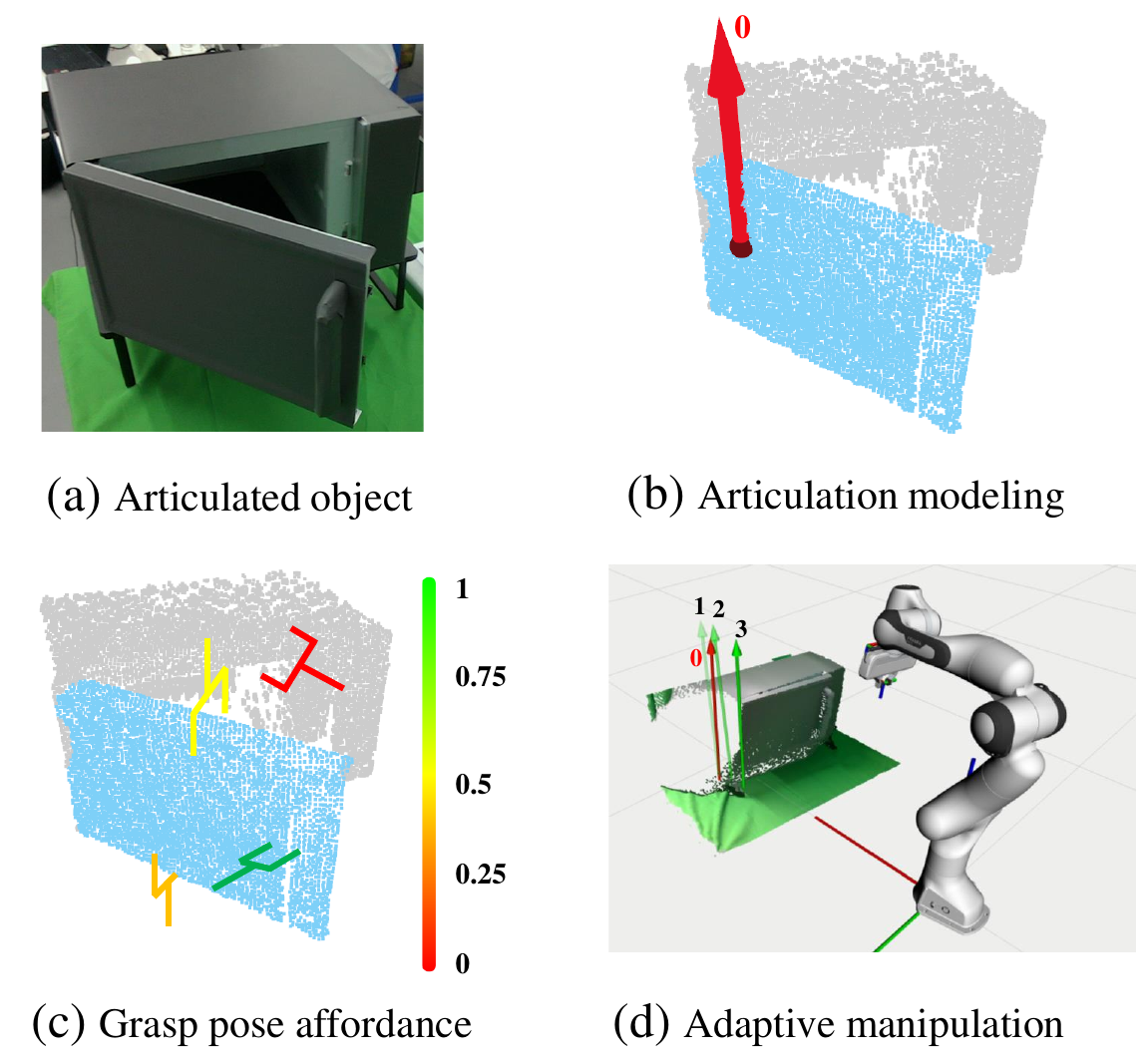}
    \caption{GAMMA framework in real microwave task. \textbf{(a)} Real microwave image to generate point clouds. \textbf{(b)} Articulated structure modeling, where blue points are segmented revolute joint and red arrow is the estimated but inaccurate joint axis and origin. \textbf{(c)} Grasp pose affordance evaluates the actionability and chooses ideal grasp poses. \textbf{(d)} Adaptive manipulation to iteratively update joint parameters. The initial joint axis(red) constantly gets closer to the ground truth after several iterations.}
    \label{Fig: teaser}
    \vspace{-5mm}
\end{figure}

Prior works focused on robot manipulation tasks by directly imitating end-to-end demonstrations~\cite{gu2022maniskill2, shen2022learning} or identifying joint parameters of specific instances or categories~\cite{li2020category}, however, they are less effective in the interaction and manipulation with novel articulated objects due to ignored geometrical and physical constraints~\cite{kaelbling2020foundation}. Despite their success, building general-purpose robots to manipulate a diverse range of articulated objects in a large variety of environments in the physical world at the human level is extremely challenging.
Therefore, recognizing and modeling the articulations are crucial for robots to generalize across-category objects with complex 3D geometries and kinematics. 

In this paper, we propose a novel, adaptive learning framework \textbf{G}eneralizable \textbf{A}rticulation \textbf{M}odeling and \textbf{M}anipulation for \textbf{A}rticulated objects (GAMMA) that leverages articulation structures model and grasp pose affordance for generalizable cross-category manipulation as depicted in Fig.~\ref{Fig: teaser}. 
Initially, GAMMA utilizes the point cloud to segment the whole object as several articulated parts and identify the physical structure and parameters of each part.
Then, GAMMA proposes a series of potential grasp poses using GraspNet~\cite{fang2020graspnet} and estimates the part-aware grasp pose affordance according to the identified articulation structure.
Finally, GAMMA guides the robot manipulation by planning optimal trajectories based on the articulation model and grasp pose affordance. GAMMA constantly improves the articulation model by re-estimating the articulation parameters based on actual trajectories and updates optimal manipulation trajectories adaptively.
In addition, GAMMA proficients at task generalization by leveraging the learned physics-informed prior of articulated objects. When transferred to unseen tasks, GAMMA can quickly recognize the articulation structure, identify joint parameters and predict grasp pose affordance from the prior knowledge, which substantially improves the success rate in cross-category articulation-associated tasks. 

In summary, the contributions of our paper are threefold:
(1) We propose the Generalizable Articulation Modeling and Manipulation for Articulated Objects (GAMMA) algorithm, which can generalize manipulation with cross-category articulated objects by estimating the articulation model and grasp pose affordance from the point cloud. 
(2) We employ physics-guided adaptive manipulation to generate articulation-feasible trajectories and iteratively estimate joint parameters to constantly improve the modeling accuracy and enhance manipulation performance. 
(3) We conduct comprehensive experiments in both simulation and real-world. Results reveal that GAMMA has strong generalizability and achieves great success in improving modeling accuracy and manipulation success rate on unseen, cross-category articulated objects.

%% file: text/related_work.tex
\section{Related Work} \label{Sec: related works}

{\bf Articulated Object Modeling.}
The development of large-scale datasets of articulated objects, such as Shape2motion~\cite{wang2019shape2motion}, PartNet-Mobility~\cite{xiang2020sapien}, and AKBNet~\cite{liu2022akb}, significantly promoted advanced research in the part segmentation and modeling of articulation structure. Part segmentation~\cite{kalogerakis20173d, yi2017syncspeccnn, lai2022stratified, geng2023gapartnet} is the preliminary of articulated object modeling and part-level manipulation, which separates different part entities of the same object category at the mask level. Early stage methods~\cite{liu2022toward, li2020category, jiang2022ditto} heuristically segment the well-explored articulated objects with a fixed number of parts. The following methods, RPM-Net~\cite{yan2020rpm} and Shape2Motion~\cite{wang2019shape2motion} leverage point-wise motion prediction to separate articulated parts from unknown objects. 
For articulation structure modeling, ANCSH~\cite{li2020category} and ReArtNet~\cite{liu2022toward} exploit densely normalized coordinate space to effectively model articulated objects with similar kinematic parameters. They achieved accurate per-part and per-joint pose estimation for unseen objects within the same category and similar kinematics but lacked generalizability for multi-tasks. 

{\bf Affordance.} 
In the manipulation tasks, affordance~\cite{gibson1977theory} indicates potential interaction modalities between the robot and objects. 
In particular, visual affordance utilizes visual information observed from objects and robots to predict the probability of successful execution of each contact pose.
Recently, extensive research has focused on learning grasping affordance~\cite{graspaffordance1, graspaffordance2, graspaffordance3, graspaffordance4, graspaffordance5} and manipulation affordance~\cite{m_affordance1, m_affordance0, m_affordance2, m_affordance3, m_affordance4} to facilitate robot-object interaction. 
For manipulation tasks with articulated objects, Adaafford~\cite{wang2022adaafford} and Where2act~\cite{mo2021where2act} utilize dense affordance maps as actionable visual representations, indicating the success rate of manipulation at each point on the 3D articulated object. 
VAT-MART~\cite{wu2021vat} proposes visual prior as representations to estimating actionable grasp pose considering geometrical constraints of articulated objects.
However, the previous methods directly learn affordance estimation without knowledge of articulation structure and have difficulty generalizing across various categories of articulated objects. 



{\bf 3D Articulated Object Manipulation.} 
The preliminary works of articulation manipulation focused on imitation learning~\cite{wong2022error, gu2023maniskill2, guhur2023instruction} that leverages demonstrations from experts to learn a manipulation policy. However, imitation learning has distribution shift problems, and collecting diverse demonstrations is time-consuming and expensive. 
In recent years, visual recognition~\cite{abbatematteo2019learning, zeng2021visual, jain2021screwnet} has emerged to estimate instance-level or category-level articulation parameters to generate manipulation trajectories. 
For instance, VAT-Mart~\cite{wu2021vat}, a pure learning-based method, employs 3D visual affordance to predict the open-loop task-specific motion trajectory of each point. 
On the contrary, a series of optimal control methods~\cite{mittal2022articulated, karayiannidis2016adaptive}, which directly use dynamic articulation models as constraints, optimize the manipulation trajectory based on recognized joint type and parameters. However, inaccurate parameter estimation may also cause repeated failure.

%% file: text/problem.tex
\section{Problem formulation}

\begin{wrapfigure}{r}{0.48\columnwidth}
  \centering
    \includegraphics[width=0.48\columnwidth, trim={3mm 3mm 3mm 2mm},clip]{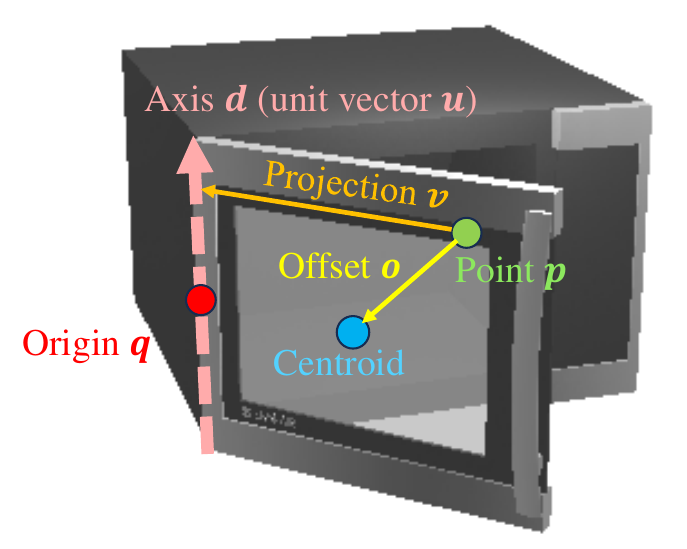}
  \caption{Illustration of points and vectors on articulated objects.}
  \label{Fig: illustration}
  \vspace{-2mm}
\end{wrapfigure}

We formulate the robot manipulation task $T$ as follows.
An unknown articulated object $M$ consists of $K$ movable parts as $M=\{m_{i}\}_{i=1}^{K}$. We observe such object $M$ via point cloud $P$ with $N$ points as $P=\{\mathbf{p}_{i}\in \mathbb{R}^{3}\}_{i=1}^{N}$.
Further, we model the object structure $J$ from the points cloud by estimating the articulation parameters $\psi_i$ for each part as $J=\{\psi_{i}\}_{i=1}^K$.
As most object only consist of one-dimensional prismatic and revolute joints~\cite{jiang2022ditto, zhang2023flowbot++, li2020category}, we can simplify the articulation parameters as $\psi_i=\{\mathbf{u}_i, \mathbf{q}_i, c_i\}$, where $\mathbf{u}_i \in \mathbb{R}^{3}$ is a unit vector of joint axis, $\mathbf{q}_i \in \mathbb{R}^{3}$ represent the origin and $c_i$ is the joint type. In addition, we formulate two variables $\mathbf{o_i} \in \mathbb{R}^{3}$ and $\mathbf{v_i} \in \mathbb{R}^{3}$ representing the vectors from point $\mathbf{p_i}$ to the centroid of articulated and to the joint axis $\mathbf{d_i}$, which are used in the articulation modeling section (Sec.~\ref{subsec: art-struc}).




%% file: text/method.tex
\section{Approach}

\begin{figure*}[t!]
  \begin{center}
   \includegraphics[width=0.9\linewidth]{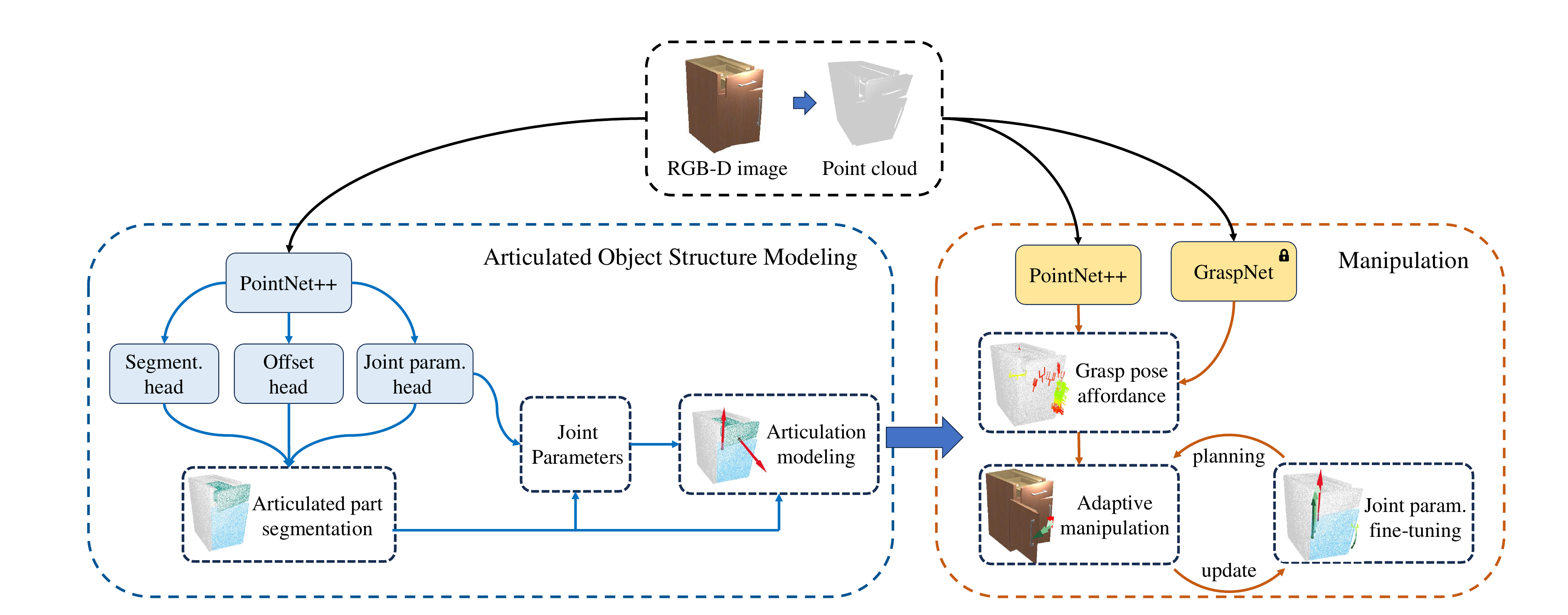}
  \end{center}
  \vspace{-2mm}
      \caption{\textbf{Pipeline of GAMMA.} We collect RGB-D images of articulated objects like a cabinet to generate point clouds. The articulation modeling block segments the articulated parts and estimates the joint parameters. The grasp pose affordance block estimates the actionability of each grasp pose and chooses the ideal ones. In the adaptive manipulation, the articulation model provides open-loop trajectory planning and we iteratively update the joint parameters with actual trajectory to improve modeling accuracy and grasping success rate.}
    \vspace{-6mm}
\label{Fig: pipeline}
\end{figure*}

We propose a novel robot manipulation algorithm, GAMMA that promotes generalizable manipulation with articulated objects. GAMMA consists of three main modules (Fig.~\ref{Fig: pipeline}): articulation modeling, part-aware grasp pose affordance and physics-guided adaptive manipulation.
First, the articulation modeling (Sec.~\ref{subsec: art-struc}) leverages articulation parameters estimated from visual observation to accurately predict kinetic constraints and improve the manipulation policy.
Then we employ part-ware grasp pose affordance (Sec.~\ref{subsec: afford}) to estimate suitable grasping poses to improve grasping quality.
Finally, the physics-guided adaptive manipulation (Sec.~\ref{subsec: phy-guid}) incorporates prior knowledge of joint parameters to acquire optimal manipulation trajectories and dynamically updates articulation parameters from executed trajectories, which constantly enhances the modeling accuracy.
In short, GAMMA aims to understand the physical structure of articulated objects to facilitate manipulation with generalized cross-category articulated objects.

\subsection{Articulation Modeling} \label{subsec: art-struc}

Understanding the structure of articulated objects helps infer the kinetic constraints of movable sections in the manipulation tasks. 
Therefore, we segment the whole object into several distinct rigid parts and estimate the articulation parameters for each part.
We first extract point-wise features from a single partial point cloud observation $P=\{\mathbf{p}_{i}\in \mathbb{R}^{3}\}_{i=1}^{N}$ with segmentation-style PointNet++~\cite{qi2017pointnet++} backbone. Then we process the raw features with segmentation, offset and joint parameters heads.

The segmentation head predicts the per-point segmentation class $\hat{c}_{i} \in \{0, 1, 2\}$ corresponding to static, revolute and prismatic parts.
The offset head generates offset vectors (Fig.~\ref{Fig: illustration}) $\hat{\mathbf{o}}_{i}\in \mathbb{R}^3$ that shift each point towards the centroid of the corresponding part.
The joint parameter head estimates the joint parameters by regressing every point $\mathbf{p}_{i}$ in the point cloud as the projection vector (Fig.~\ref{Fig: illustration}) of each point to the joint axis $\hat{\mathbf{v}}_{i}\in \mathbb{R}^3$ and the estimated joint axis direction $\hat{\mathbf{d}}_{i}\in \mathbb{R}^3$ to the axis $\mathbf{u}$.

Finally, we jointly optimize the losses of segmentation, offset, and joint parameters as,
\begin{equation}\label{deqn_ex1}
\begin{aligned}
\mathcal{L} = \frac{1}{N} \sum_{i}^N & \Bigl[ \mathcal{L}_{c}(\hat{c}_{i},c_{i}) + \mathcal{L}_{o}(\hat{\mathbf{o}}_{i},\mathbf{o}_{i}) \\
& + \mathcal{L}_{v}(\hat{\mathbf{v}}_{i},\mathbf{v}_{i}) + \mathcal{L}_{d}(\hat{\mathbf{d}}_{i},\mathbf{d}_{i}) \Bigr],
\end{aligned}
\end{equation}
where, ${c}_{i}$, $\mathbf{o}_{i}$, $\mathbf{v}_{i}$, and $\mathbf{d}_{i}$ are the ground-truth values of $\hat{c}_{i}$, $\hat{\mathbf{o}}_{i}$, $\hat{\mathbf{v}}_{i}$, and $\hat{\mathbf{d}}_{i}$. 
$\mathcal{L}_{c}(\hat{c}_{i},c_{i})$ denotes the focal loss to balance semantic distribution~\cite{lin2017focal}; $\mathcal{L}_{o}(\hat{\mathbf{o}}_{i},\mathbf{o}_{i})$ and $\mathcal{L}_{v}(\hat{\mathbf{v}}_{i},\mathbf{v}_{i})$ impose constraints on both the L1 distance and direction of the point offsets; $\mathcal{L}_{d}(\hat{\mathbf{d}}_{i},\mathbf{d}_{i})$ optimize the axis direction estimation. In detail, offset loss is as
\begin{equation} \label{deqn_ex2}
\mathcal{L}_{o}(\hat{\mathbf{o}}_{i},\mathbf{o}_{i}) = \|\hat{\mathbf{o}}_{i}-\mathbf{o}_{i}\| - (\frac{\mathbf{o}_{i}}{\|\mathbf{o}_{i}\|}_{2}\cdot \frac{\hat{\mathbf{o}}_{i}}{\|\hat{\mathbf{o}}_{i}\|}_{2})
\end{equation}

By minimizing the loss function, we primarily model the articulated object with predicted segmentation $\hat{c}_{i}$, offset $\hat{\mathbf{o}}_{i}$, point projection $\hat{\mathbf{v}}_{i}$ and joint axis estimation $\hat{\mathbf{d}}_{i}$. 

After completing the training phase, we freeze all parameters during feature inference. Then, we perform semantics-aware and axis-aware clustering to group the points into separate clusters. Specifically, the points of revolute or prismatic parts are first selected by the predicted semantics label. Then, they are shifted by offset vectors to form a more compact 3D distribution $\{(\mathbf{p}_{i} + \hat{\mathbf{o}}_{i})\}$ where the intra-part points are spatially closer, and they are shifted by projection vector to form a more compact 3D distribution $\{(\mathbf{p}_{i} + \hat{\mathbf{v}}_{i})\}$ where the intra-part points are spatially linear, involving the projection of points onto a common axis. Considering the density of feature sets $\{(\mathbf{p}_{i} + \hat{\mathbf{o}}_{i}), (\mathbf{p}_{i} + \hat{\mathbf{v}}_{i})\}$, we adopt DBSCAN~\cite{ester1996density} to group these points into a part cluster as $M=\{m_{i}\}_{i=1}^{K}$. For each segmented part, we utilize the estimated per-point projection vector and axis direction to vote the joint parameters.

\subsection{Part-Aware Grasp Pose Affordance} \label{subsec: afford}
We extend the articulation model with grasp pose affordance, which interprets the high-dimensional visual information as practical interactive positions on the objects to facilitate robot manipulation. In a specific manipulation task $T$, we tackle each articulated part with the following steps.

First, we generate a series of grasp poses using GraspNet~\cite{fang2020graspnet}. Secondly, we concatenate features of global geometrical feature $f_{gg}$ from PointNet++ backbone~\cite{qi2017pointnet++}, articulation parameter feature $f_{ap}$, and the contact pose feature $f_{gp}$ from an MLP. Finally, we utilize the concatenated features to predict the actionability score $s_{g|P,\psi} \in [0, 1]$ to evaluate the feasibility of each grasp pose.

\subsection{Physics-Guided Adaptive Manipulation} \label{subsec: phy-guid}

In the previous sections, we built the articulation model and generated grasp pose affordance from point cloud to provide kinetic constraint in open-loop planning. However, the uncertainty and errors in observation make the raw model inaccurate in the manipulation. 
To address this issue, we employ physics-guided adaptive manipulation to iteratively update the articulation model by minimizing the errors between planned and actual trajectories. This process helps improve model accuracy in diverse manipulation tasks with unseen articulation objects. In the following section, we detail the trajectory generation method for revolute and prismatic joints and parameter adjustment.

In the physics-guided trajectory planning, a \emph{revolute joint} is defined by the joint axis $\mathbf{u}$ and origin $\mathbf{q}$. The trajectory of contact point $\mathbf{p}$ with respect to a rotation angle $\theta$ is
\begin{equation} \label{deqn_ex3}
\mathcal{P} = \cos(\theta)\mathbf{I} \cdot \mathbf{p} + (1 - \cos(\theta))\mathbf{u}\mathbf{u}^T \cdot \mathbf{p} + \sin(\theta) \mathbf{R} \cdot \mathbf{p} + \mathbf{q}
\end{equation}
where $\mathbf{I}$ denotes an identity matrix and $\mathbf{R}$ denotes the skew-symmetric matrix of $\mathbf{u}$. Therefore, we can plan a feasible trajectory with rotation angle $\theta$ from the initial value to the target.
Different from the revolute joint, the \emph{prismatic joint} only translates along the joint axis $\mathbf{u}$. We can obtain the trajectory of contact points $\mathbf{p}$ with varying translation distance $\delta$ as,
\begin{equation}
\label{deqn_ex4}
\mathcal{P} = \mathbf{p} + \delta \mathbf{u}.
\end{equation}


We plan the trajectory of grasp points in $L$ steps in a manipulation task $T$ with articulation parameters $\psi$ at time step $t$ as $\tau_{t}^{\text{plan}}=\{\mathcal{P}_{i}^{\text{plan}}\}_{i=1}^{L}$. 
In the actual manipulation, we adopt receding horizon control and only execute the first $H$ ($H<L$) steps of the planned $\tau_{t}^{\text{plan}}$. As a result, the actual trajectory sampled from the real robot is $\tau_{t}^{\text{actual}}=\{\mathcal{P}_{i}^{\text{actual}}\}_{i=1}^{H}$. We finally optimize the articulation parameters $\psi$ by minimizing the product of matching matrix $C_t\in\mathbb{R}^{H \times L}$ in Hungarian algorithm~\cite{kuhn1955hungarian} and the errors between planned and actual trajectories as,
\begin{equation}
\label{deqn_ex5}
\mathcal{L}_{t}(\psi) = \frac{1}{H} \sum_{h=0}^{H}\sum_{l=0}^{L} \|\mathcal{P}_{h}^{\text{actual}} - \mathcal{P}_{l}^{\text{plan}} \| C_t[h, l].
\end{equation}
By iteratively planning trajectories and optimizing articulation parameters with actual trajectories sampled from the real world, we can constantly improve the modeling accuracy of the articulated object.

%% file: text/experiment.tex
\section{Experiments}

In this section, we conduct comprehensive robot manipulation tasks with articulated objects in both simulated and real-world environments. We compare the performance of GAMMA with other baselines to answer the following question: 
1) Can GAMMA effectively model articulated structure from point cloud by segmenting the articulation parts? 
2) Can physics-guided adaptive manipulation 
 based on grasp affordance improve the model accuracy and manipulation performance? 
3) Can GAMMA effectively generalize skills learned from perception prior on unseen and cross-category articulation in both simulated and real-world environments?

\renewcommand\arraystretch{1.2}
\begin{table*}[ht!]
\caption{Articulation Modeling Result}
\vspace{-3mm}
\begin{center}

\begin{threeparttable}
\resizebox{\linewidth}{!}{
\begin{tabular}{c|l|cc|cc|cc|cc}
\toprule 

\multirow{2}{*}{~} & \multirow{2}{*}{Category} & \multicolumn{2}{c|}{AP75$^2$($\%$) $\uparrow$} & \multicolumn{2}{c|}{Type Acc.$^2$($\%$) $\uparrow$} & \multicolumn{2}{c|}{Axis error$^2$($^\circ$) $\downarrow$} & \multicolumn{2}{c}{Origin error$^2$(cm) $\downarrow$} \\

& & ANCSH~\cite{li2020category} & GAMMA & ANCSH & GAMMA & ANCSH & GAMMA & ANCSH & GAMMA \\

\hline \hline

\multirow{5}{*}{Unseen instances$^1$}
  & Cabinet & 66.4& \textbf{80.9} & 95.8 & \textbf{98.9} & 19.95 & \textbf{9.48} & 37.24 & \textbf{15.63} \\
 & Door & 60.3 & \textbf{97.1} & \textbf{100} & \textbf{100} & 19.45 & \textbf{2.27} & \textbf{4.90} & 10.72  \\
 & Microwave & \textbf{100} & \textbf{100} & \textbf{100} & \textbf{100} & 4.55 & \textbf{0.87} & 6.29 & \textbf{2.62}\\
 & Refrigerator & 52.9 & \textbf{98.3} & \textbf{100} & \textbf{100} & 19.33 & \textbf{6.99} & 22.2 & \textbf{1.69} \\
&\emph{Average} & 69.9 & \textbf{94.1} & 98.95 & \textbf{99.7} & 15.82 & \textbf{4.90} & 17.66 & \textbf{7.66} \\
\midrule
\multirow{4}{*}{Unseen categories$^1$}
 & Safe & \textbf{100} & \textbf{100} & 100 & \textbf{100} & 10.53 & \textbf{3.75} & 8.09 & \textbf{1.36} \\
 & Washing machine & 81.8 & \textbf{90.2} & \textbf{100} & \textbf{100} & 24.71 & \textbf{7.77} & 6.24& \textbf{4.54} \\
 & Table &16.1 & \textbf{50.5} & 56.9 & \textbf{82.0} & 45.25 & \textbf{15.04} & 34.57& \textbf{18.69} \\
&\emph{Average} & 66.0 & \textbf{80.2} & 85.6 & \textbf{94.0} & 26.83 & \textbf{8.85} & 16.3 & \textbf{8.20} \\
\bottomrule
\end{tabular}}

\begin{tablenotes} 
\item[1] \textbf{Unseen instances} include data from four categories of objects used in the model training, which are used to validate and avoid overfitting. \textbf{Unseen categories} include data from three totally new categories, which are used to test the generalizability of models.
\item[$^2$] Four evaluation metrics are \textbf{AP75}: average precision under IoU 0.75 of instance part segmentation; \textbf{Type Acc.}: joint type classification accuracy; \textbf{Axis error}: joint axis error; and \textbf{Origin error}: joint origin error.
\end{tablenotes}

\end{threeparttable}
\end{center}
\label{tab:model}
\vspace{-6mm}
\end{table*}

\subsection{Experimental Setup}

{\bf Environments.}
We conduct simulation experiments in the SAPIEN simulator~\cite{xiang2020sapien}. SAPIEN simulator provides physical simulation for robots, rigid bodies, and articulated objects, which also has photo-realistic rendering that facilitates sim-to-real generalization. We train GAMMA with 4 categories of objects: cabinet (revolute and prismatic joints), door (revolute joint), refrigerator (revolute joint), and microwave (revolute joint). We evaluate the performance and generalization ability of GAMMA in 3 unseen categories: safe (revolute joint), table (revolute and prismatic joints), and washing machine (revolute joint). In addition, we also set up real-world experiments with a 7-Dof Franka robot. We first mount an RGB-D camera RealSense L515 on the robot's wrist to sense point clouds of all articulated objects. Then we apply GAMMA to manipulate two unseen objects: cabinet (revolute and prismatic joints) and microwave (revolute joint). 

{\bf Datasets.}
We train GAMMA with PartNet-Mobility dataset~\cite{wu2021vat, mo2021where2act}, which has 562 different articulated objects in 7 categories. In detail, each articulated object contains one or more prismatic or revolute rigid parts. The initial joint states are uniformly distributed within the joint limit range. 
For observation, we used an RGB-D camera with $448\times448$ resolution to spherically sample camera viewpoints in front of the target object with yaw angle in $[-90^{\circ}, 90^{\circ}]$ and pitch angle in  $[30^{\circ}, 60^{\circ}]$.
In practice, we sample 20 different states for each object and 5 viewpoints for each state to generate point clouds. In total, we generate $32800$, $9400$, and $14000$ samples for training, validation, and testing, respectively, using $328$, $94$, and $140$ objects. 

\begin{figure}[th]
    \centering
    \includegraphics[width=0.95\columnwidth]{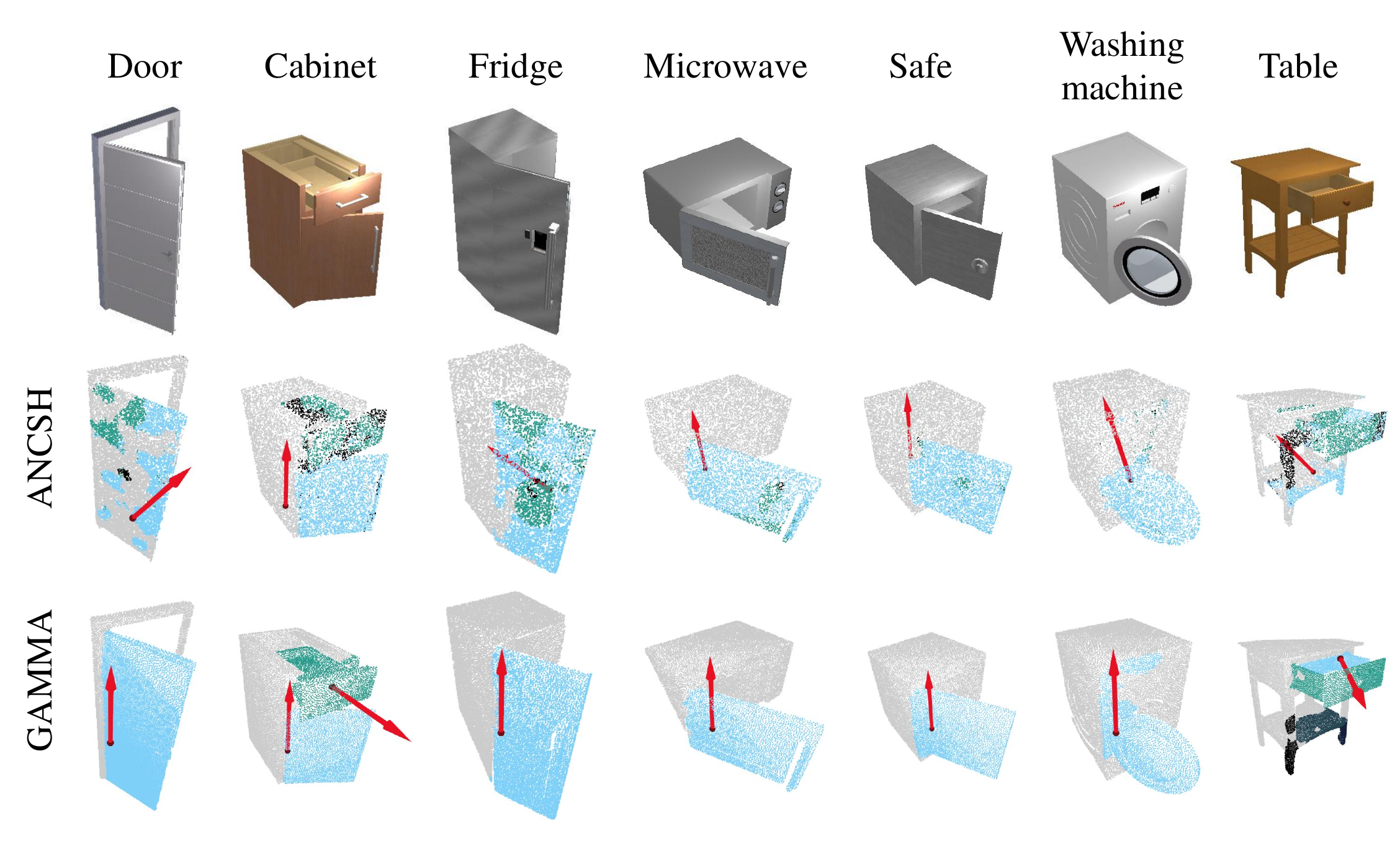}
    \caption{We implement ANCSH and GAMMA to model articulated objects in 7 categories. The first rows are images in the simulation environment. The second and third rows are segmentation results of articulated parts, marked as blue, green and dark green points. Each color represents a separate modeled articulated part. The red arrow and dot denote the estimated joint axis direction and origin position.}
    \label{Fig: objects}
    \vspace{-5mm}
\end{figure}

\noindent {\bf Baselines of Articulation Modeling and Manipulation.}
\begin{itemize}[left=0pt]
\item ANCSH~\cite{li2020category} is a state-of-the-art articulated object modeling method. It first segments the object as articulated parts with single-view point clouds; then transforms the points into a normalized coordinate space to estimate joint parameters.
\item Reinforcement learning (RL, TD3~\cite{fujimoto2018addressing}) is a widespread baseline of robot manipulation tasks. The observation includes point clouds and end-effector states, and the action is the incremental changes of the end-effector's state.

\item Where2Act~\cite{mo2021where2act} selects grasping points with better actionability for manipulation and generates short-term manipulation actions (pushing or pulling) on each point. We test Where2Act in the long-term manipulation tasks by repeatedly executing selected actions multiple times to progressively manipulate the articulated parts.

\item VAT-Mart~\cite{wu2021vat} employs 3D object-centric actionable visual priors for manipulation tasks. This model predicts interaction-aware and task-aware visual action affordance and trajectory proposal for manipulation tasks.
\end{itemize}

\subsection{Articulated Object Modeling Results}

{\bf Tasks and Metrics.} In the articulated object modeling task, we evaluate the performance of the articulation model from point clouds. We first train ANCSH~\cite{li2020category} and GAMMA with 4 training categories in PartNet-Mobility dataset~\cite{wu2021vat, mo2021where2act} and evaluate in 3 unseen categories. Finally, we compare three modeling accuracy metrics:
average precision under IoU 0.75 of instance part segmentation, joint type classification accuracy, joint axis error, and joint origin error. 

{\bf Results.} Fig.~\ref{Fig: objects} and Table~\ref{tab:model} presents the articulation modeling results of ANCSH and GAMMA. Results show that both ANCSH and GAMMA have high joint-type prediction accuracy in most categories. However, ANCSH cannot easily recognize tables because they are unseen and have extra stand part that interferes with the articulation structure. In addition, GAMMA significantly outperforms ANCSH in segmenting articulated parts. 
The reason is GAMMA not only abstracts visual representation from point cloud as ANCSH does, but also clusters points with the same semantics by shifting them along the axis direction and projecting them to a more compact and information-rich feature set. Therefore, GAMMA can easily segment the articulated parts since their features are more distinguishable in the transformed vector space. 

In addition, although ANCSH has shown success in estimating joint parameters on seen articulated objects, it failed to estimate joint axis direction in the cross-category tasks and has larger origin errors than GAMMA in the unseen categories (Table~\ref{tab:model}). In the experiments, we find a potential reason that ANCSH transforms all points to a normalized coordinate space to estimate joint parameters. This approach can work when all objects have similar sizes and structures, but in the cross-categories tasks, where the objects have diverse sizes and structures, the points normalized coordinate space might entwine heavily and lead to totally wrong axis estimation. On the contrary, GAMMA does not apply normalized coordinate space but utilizes all projected points and estimated axis direction to vote for the final joint parameters. This method leverages the geometrical structure of articulated objects and easily generalizes to unseen cross-category tasks.

\subsection{Articulation Manipulation Results}

\renewcommand\arraystretch{1.3}
\begin{table*}[th]
\caption{Articulated Object Manipulation Results}
\vspace{-3mm}
\begin{center}
\begin{threeparttable}

\begin{tabular}{c|cccccccc}
\toprule
    \multirow{3}{*}{~} & \multicolumn{4}{c|}{Unseen instances$^1$} & \multicolumn{4}{c}{Unseen categories$^1$} \\
    & \multicolumn{2}{c}{door} & \multicolumn{2}{c|}{drawer} & \multicolumn{2}{c}{door} & \multicolumn{2}{c}{drawer} \\
    ~ & pushing & pulling & pushing & \multicolumn{1}{c|}{pulling}  & pushing & pulling & pushing & pulling \\
    \hline  \hline
 
RL(TD3)~\cite{fujimoto2018addressing}  & 5.63 & 1.20 & 5.27 & \multicolumn{1}{c|}{3.76} &3.91 & 0.79 & 2.53 & 3.08 \\
Where2Act~\cite{mo2021where2act} & 31.59 &	7.53 &	30.58 &	\multicolumn{1}{c|}{9.30} & 34.52 &	5.04 &	22.69 &	 8.61\\

VAT-Mart~\cite{wu2021vat}  &53.84 & 14.79 & 55.70 & \multicolumn{1}{c|}{41.08}& 38.06 & 12.94 & 36.74 & 29.07 \\

GAMMA(w/o afford.)$^2$ &51.79 & 35.05 & 54.46 & \multicolumn{1}{c|}{54.55} & 59.26 & 53.66 & 46.20 & 35.62\\

GAMMA(w/o adpt.)$^2$ & 69.62 & 41.62 & 61.29 & \multicolumn{1}{c|}{42.86} & \textbf{87.41} & 58.15 & 63.49 & 38.10 \\

GAMMA (ours) & \textbf{72.31} & \textbf{48.06} & \textbf{68.80} & \multicolumn{1}{c|}{\textbf{57.60}} & 86.86 & \textbf{63.89} & \textbf{65.09} & \textbf{42.11} \\

\bottomrule
\end{tabular}

\begin{tablenotes}
    \item[1] Unseen instances and unseen categories have the same definition in Table~\ref{tab:model}. We perform experiments on each object with 5 trials and calculate the average \textbf{success rate ($\%$)} in every task.
    \item[2] GAMMA(w/o afford.) and GAMMA(w/o adapt.) are two ablation studies that remove grasp pose affordance and adaptive manipulation from the GAMMA framework.
\end{tablenotes}

\end{threeparttable}
\end{center}
\label{tab: manipulation}
\vspace{-5mm}
\end{table*}

{\bf Tasks and Metrics.} We test 4 articulation manipulation tasks in Where2Act~\cite{mo2021where2act} and VAT-Mart~\cite{wu2021vat}: pushing door, pulling door, pushing drawer and pulling drawer on both seen and unseen objects. Specifically, the pulling tasks are harder than pushing because they require accurate grasping of the handles or graspable edges. In addition, we especially emphasize the generalizability of manipulating cross-category articulated objects, so we extensively evaluate 4 tasks on $94$ unseen objects with the same categories of training set and $140$ objects in the unseen categories.
Finally, we compare the success rate of 4 basic methods: RL(TD3)~\cite{fujimoto2018addressing}, Where2Act~\cite{mo2021where2act}, VAT-Mart~\cite{wu2021vat}, GAMMA(ours) and two ablation methods: GAMMA without adaptive manipulation (w/o adpt.), GAMMA without grasp pose affordance (w/o afford.). 


{\bf Results.} 
We present the success rate of manipulation results in Table.~\ref{tab: manipulation}. 
RL(TD3) baseline learns from scratch and only achieves little success. Where2Act slightly improves the success rate but it lacks long-horizon trajectory planning. VAT-Mart, on the other hand, substantially increases the success rate on all tasks by estimating visual action affordance and predicting trajectory proposals. As a comparison, we find GAMMA significantly outperforms all baselines and improves the success rate significantly which indicate that the cross-category performance of each module contributes to a significant improvement in the overall robotic manipulation.

In addition, we implement two ablation studies to analyze the influence of grasp pose affordance and adaptive manipulation. In the GAMMA without affordance (w/o afford) experiment, we randomly chose a grasp pose on the articulated object. The performance degrades to much lower success rates because the edges of articulated objects are not actionable. In the GAMMA without adaptive manipulation (w/o adapt.) experiment, we only apply open-loop trajectory planning to manipulate the articulated objects. The ablation causes obvious decreases in success rate in unseen instances and moderate or no decreases in the unseen categories. These two ablations reveal that grasp pose affordance and adaptive manipulation crucially contribute to the outstanding performance of the GAMMA algorithm.

\subsection{Real-World Experiments}
We apply our method to real-world objects to verify
its generalization ability. In the real world, we set up 20 different camera viewpoints, five states for each object, and generated 100 point clouds for each object, totaling 200 point clouds. To reduce the gap between simulated and real point cloud data, we randomly select $10\%$ to $30\%$ of the points in the simulated point cloud and add Gaussian noise with a mean of $0$ and a standard deviation of $0.03$. 

\begin{figure}[th]
\centering\includegraphics[width=0.95\columnwidth]{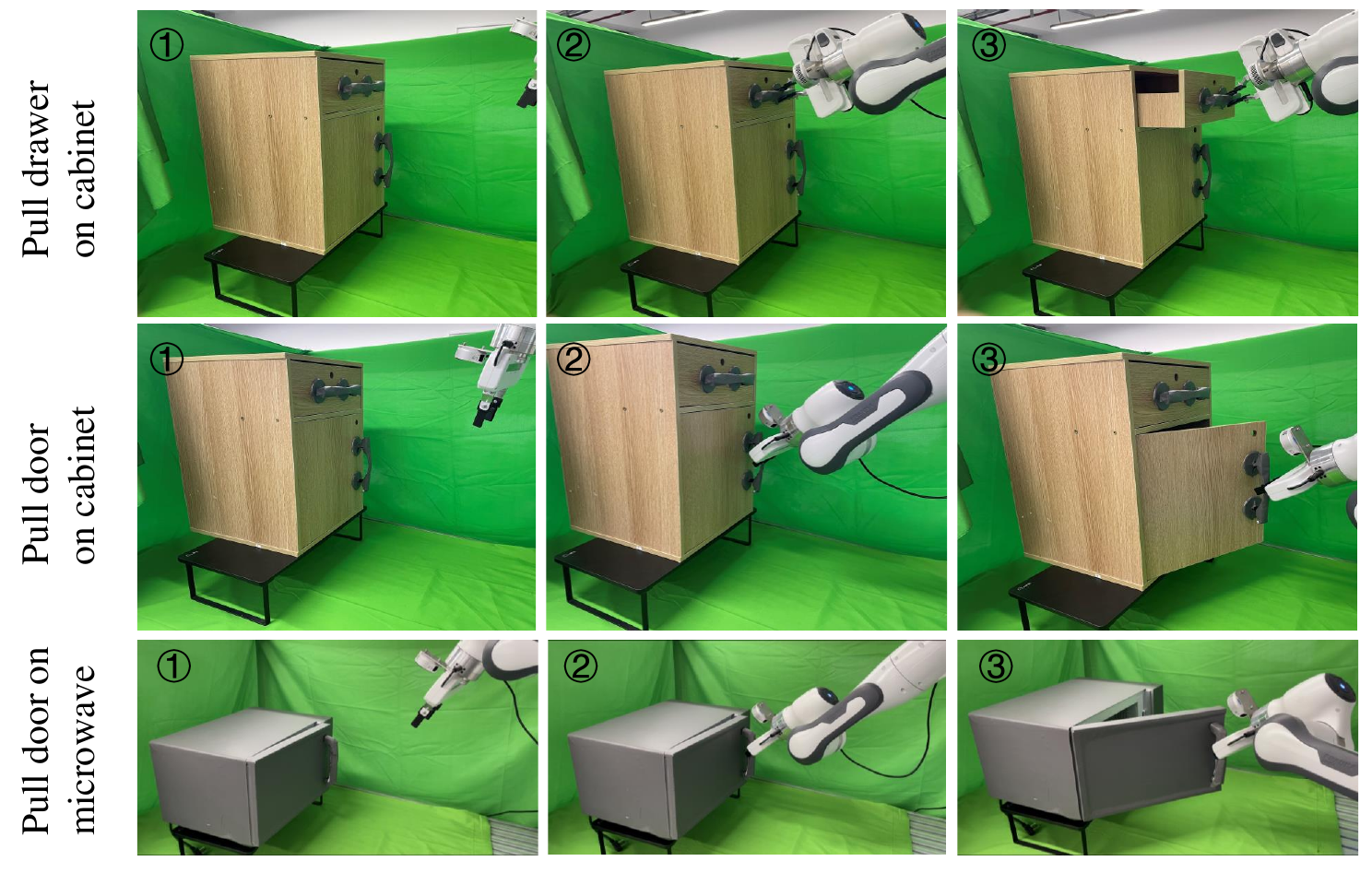}
    \caption{We implement GAMMA in the real-world experiments. Manipulation tasks include pulling the drawer and door on a cabinet and pulling the door on a microwave.}
    \label{Fig: real tasks}
    \vspace{-1mm}
\end{figure}

First, we use the model trained on simulated data to predict real-world point clouds for both the microwave and cabinet. 
The microwave has an average axis error of $6.14^{\circ}$ and an origin error of $7.36 cm$. For the cabinet, the door has an average axis error of $5.33^{\circ}$and an origin error of $7.7 cm$, while the drawer has an axis error of $6.21^{\circ}$. More results can be found on supplementary materials. The results demonstrate our proposed method can generalize to real-world point clouds well. 
Then we estimate the grasp pose affordance and apply adaptive manipulation on three robotic manipulation tasks of pulling the door and drawer on a cabinet and the door on a microwave. The effectiveness of overall framework has been verified through real-world robotic manipulation as shown in Fig.~\ref{Fig: real tasks} and supplementary video.

%% file: text/conclusion.tex
\section{Conclusion}

In this paper, we propose the generalized articulation modeling and manipulation (GAMMA) framework that estimates articulation parameters and grasp pose affordance from point clouds. In addition, GAMMA utilizes actual trajectory to iteratively update articulation parameters to improve manipulation performance. Experiments show that GAMMA significantly outperforms baselines in both articulated object modeling and manipulation, and has outstanding generalizability in cross-category tasks.

%% file: text/acknowledgments.tex
\section{Ackonwledgments}
This work was supported by the National Key Research and Development Project of China (No.2022ZD0160102, No.2021ZD0110704), Shanghai Artificial Intelligence Laboratory, XPLORER PRIZE grants, and National Natural Science Foundation of China (62302143, 52305030).